\documentclass[journal]{vgtc}                
\ifpdf
  \pdfoutput=1\relax                   
  \usepackage{graphicx}                
  \DeclareGraphicsExtensions{.pdf,.png,.jpg,.jpeg} 
\else
  \usepackage{graphicx}                
  \DeclareGraphicsExtensions{.eps}     
\fi%

\graphicspath{{figures/}{pictures/}{images/}{./}} 

\usepackage{microtype}                 
\PassOptionsToPackage{warn}{textcomp}  
\usepackage{textcomp}                  
\usepackage{mathptmx}                  
\usepackage{times}                     
\usepackage{cite}                      
\usepackage{tabu}                      
\usepackage{booktabs}                  
\usepackage{algorithm}
\usepackage{algorithmic}
\usepackage{fontawesome}
\usepackage{hyperref}
\usepackage{amsfonts}
\usepackage{multirow}
\usepackage{makecell}
\usepackage{amsmath}

\newcommand{\sysname}{IDLat} 

\onlineid{1018}

\vgtccategory{algorithm/technique}

\title{{\sysname}: An Importance-Driven Latent Generation Method for \\ Scientific Data}

\author{Jingyi~Shen, 
        Haoyu~Li, 
        Jiayi~Xu, 
        Ayan~Biswas, 
        and~Han-Wei~Shen, \textit{Member,~IEEE}}
\authorfooter{
\item
 Jingyi Shen, Haoyu Li, Jiayi Xu and Han-Wei Shen are with the Department of Computer Science and Engineering, The Ohio State University. E-mail: \{shen.1250, li.8460, xu.2205 and shen.94\}@osu.edu.
\item
 Ayan Biswas is with Los Alamos National Laboratory. E-mail: ayan@lanl.gov.
}

\shortauthortitle{Shen \MakeLowercase{\textit{et al.}}: IDLat: An Importance-Driven Latent Generation Method for Scientific Data}

\abstract{ 
Deep learning based latent representations have been widely used for numerous scientific visualization applications such as isosurface similarity analysis, volume rendering, flow field synthesis, and data reduction, just to name a few. 
However, existing latent representations are mostly generated from raw data in an unsupervised manner, which makes it difficult to incorporate domain interest to control the size of the latent representations and the quality of the reconstructed data. 
In this paper, we present a novel importance-driven latent representation to facilitate domain-interest-guided scientific data visualization and analysis. 
We utilize spatial importance maps to represent various scientific interests and take them as the input to a feature transformation network to guide latent generation. We further reduced the latent size by a lossless entropy encoding algorithm trained together with the autoencoder, improving the storage and memory efficiency. We qualitatively and quantitatively evaluate the effectiveness and efficiency of latent representations generated by our method with data from multiple scientific visualization applications. 
} 


\CCScatlist{ 
 \CCScat{K.6.1}{Management of Computing and Information Systems}%
{Project and People Management}{Life Cycle};
 \CCScat{K.7.m}{The Computing Profession}{Miscellaneous}{Ethics}
}

\vgtcinsertpkg

\begin{document}



\maketitle
\section{Introduction}
As machine learning techniques become increasingly more ubiquitous for scientific visualization and analysis, 
latent representations generated by autoencoders have attracted great attentions of researchers in recent years. Latent representations have been successfully demonstrated to retain essential information in the original data, and can be used for similarity analysis~\cite{Han2020FlowNet, Porter2019TSselectionTVM, Dai2021IsoExplorerAI, li2022local,cheng2018deep}, generation of visualizations~\cite{berger2018GanVR}, synthesis of simulations~\cite{wiewel2019latentphysics, kim2019deepfluids, wiewel2020latentsubdivision}, data reductions~\cite{Zhang2021Multibranch, liu2021AESZ}, and have been applied to multivariate volumetric data~\cite{Porter2019TSselectionTVM}, streamlines and stream surfaces~\cite{Han2020FlowNet},  isosurfaces~\cite{Dai2021IsoExplorerAI}, and particles~\cite{li2022local}. 

Although latent representations for large-scale scientific data have been used extensively, there are still several challenges. First, domain scientists have diverse interests in different data portions, but latent representations trained using unsupervised approaches have limited support for incorporating such domain interests.
Given that scientific data complexity varies across space and time~\cite{fox2018feature}, domain scientists' interests should be taken into account during latent generation so that it is possible to perform importance-driven scientific data explorations as well as to reduce data that are not deemed important. To the best of our knowledge, related works only support generating latent representations associated with simulation parameters~\cite{kim2019deepfluids, wiewel2020latentsubdivision}, time~\cite{ wiewel2019latentphysics}, and aggregated queries~\cite{Wang2021NeuralCubes}. 
Second, how to represent diverse domain interests in a unified way for latent generation is non-trivial. Domain interest in scientific visualization can be defined in many ways, either mathematically related to physical attributes or spatially/temporally related to particular ranges~\cite{Viola2005feature, Viola2006FocusAttention, gosink2010multivariateQDV, wang2009AppDrCompTV, viola2004imporVR}. A generalized representation is required to incorporate different types of scientific interests. 
Third, the costs of importance-driven latent generation can be high. Previous latent representations are tightly coupled with specific scientific visualization applications~\cite{Porter2019TSselectionTVM, Dai2021IsoExplorerAI, Han2020FlowNet}. If scientists change their interests during exploration, re-training the model will be needed but can be prohibitively expensive. Also, current latent generation methods cannot adapt the size of the latent to the domain interest once neural network architecture is determined, leading to high storage and I/O costs.


In this paper, to generate latent representations guided by scientific interests, we propose an \textbf{I}mportance-\textbf{D}riven \textbf{Lat}ent generation method (\textbf{\sysname}) based on a convolutional autoencoder to combine the power of the convolution operations for extracting local features and the autoencoder for representation learning~\cite{bengio2013representationLearn}.
First, to incorporate domain interests into latent representations, we extend the basic autoencoder with a feature transformation network that takes domain interest as an input to guide the mapping from scientific data to latent representations. 
Second, based on the proposed network, we represent various types of domain interests with discretized spatial importance maps. Every element in the importance map is a real value indicating how vital this spatial location is when generating the latent representation. The importance values can be derived mathematically based on the domain or heuristically based on distances, distributions, locations, etc., depending on the underlying scientific applications. With the location-wise control of spatial importance, we can flexibly represent various types of scientific interests and use them to guide latent generation. 
Third, our model only needs to be trained once for each dataset, and used even when scientists change the definition of importance. The produced latents are optimized in storage size with the help of feature transformation networks and a lossless entropy encoding module. 
The motivation for jointly pursuing importance-driven latent and compression can be summarized into two aspects. The first is to further reduce the storage cost of scientific data based on its importance. Latent representations are compact, but their sizes are determined by the network architecture, not the amount of information according to domain interest. To optimize the usage of storage, we quantize and compress latent with importance taken into account, i.e., reduce the size of latent for unimportant data. 
The second is to improve the effectiveness of latent in representing scientific features. The original data may contain unimportant information such as noise or non-feature regions which compromise latent’s ability to represent features. However, with importance control and entropy constraint in the latent space, the model will optimize the utilization of limited latent dimensions by preserving more important information and sacrificing the unimportant information.  
As a result, each latent is instructed to encode important information effectively.

Our latent generation workflow is as follows. First, spatial importance maps are generated based on scientific interests. Second, both the original scientific data and importance maps are taken as input to our model, which produces latent representations controlled by the importance map. Third, we quantize the generated latents into discrete symbols. Fourth, given that the entropy of discrete latents will be different under different importance settings, we apply lossless entropy encoding on the discrete latent vectors to further reduce the latent size. 
After the model is trained, we support visualization and analysis in both latent space and data space. In latent space, the discrete latent representations are losslessly recovered through entropy decoding for scientific analysis such as similarity comparison and feature exploration. In data space, the discrete symbols are further decoded to obtain reconstructed data through the autoencoder's decoder.

Our latent representation is useful for scientific visualization and analysis due to its compactness and effectiveness in preserving domain interests. 
Each latent representation is forced to focus on representing data of interest instead of all details of the raw data, which amplifies the more salient information and reduces the effect of noise, resulting in more salient and robust data representations. 
By transforming data into compact latent representations, similarity comparison or distance computation between data becomes efficient and robust. 
Also, it reduces the storage cost by only saving the compressed latent representations that can later be used for downstream scientific analysis tasks such as projection, retrieval, feature exploration, clustering, query, etc.

We qualitatively and quantitatively evaluate the usefulness and effectiveness of our importance-driven latent representations through data reconstruction and latent space exploration tasks on three scientific datasets. 
In summary, the contributions of our work are threefold:  
\vspace{-5pt}
\begin{itemize}
\item First, we present a novel and flexible pipeline for generating importance-driven scientific data representation with an autoencoder model.
\vspace{-6pt}
\item Second, we utilize a location-based importance map to incorporate domain interests into the generation of latent representation.
\vspace{-14pt}
\item Third, we further reduce the size of latent representation through entropy encoding to reduce the I/O and storage costs. 
\end{itemize}

\section{Related Works}
Our study makes use of a deep learning based latent representation for importance-driven data visualization and analysis. We summarize the related works of these two fields. 

\subsection{Latent Representations in Scientific Visualization} \label{LatentSciVis}
In scientific visualization, there are three main usages of autoencoders. 
The first, and the most related one to our work, is to use autoencoders for user-controlled data synthesis. 
Wiewel et al.\cite{wiewel2020latentsubdivision} converted raw volume data into latent representations and controlled data properties such as velocity or density through different latent dimensions. 
Berger et al. \cite{berger2018GanVR} proposed to learn the mapping from transfer functions to rendered volumes with an encoder-decoder architecture. By traversing the latent space and generating rendered images of the volume under various viewpoints and transfer functions, scientists can get a better understanding of the volume features efficiently.
Kim et al. \cite{kim2019deepfluids} proposed a latent space integration network to learn the mapping of latent representations from the current time step to the next time step. 
Second, latent representations are also used as feature descriptors of the raw data to select representatives. 
FlowNet \cite{Han2020FlowNet} proposed to identify representative flow lines or surfaces in the lower dimensional latent space by applying density-based clustering on latent representations. 
To select representative time steps for volumetric time-varying data, instead of using handcrafted features, Porter et al. \cite{Porter2019TSselectionTVM} adopted autoencoders to learn a representation for each volume and selected representations in the t-SNE projection.
The third usage of autoencoders is data reduction. AE-SZ~\cite{liu2021AESZ} and multi-branch decoder network \cite{Zhang2021Multibranch} demonstrate the effectiveness of autoencoders for scientific data reduction. 

However, existing autoencoder-based works assume every data element is equally important without considering scientists' interest when generating the latent. Also, from a data reduction point of view, knowing which region scientists have low interests and thus can afford to have a lower quality will help achieve a better trade-off between the size and the quality of the latent representation. Therefore, we extend the basic autoencoder into one conditioned on user interest. 

\subsection{Importance-Driven Visualization and Analysis} 
For different scientific applications, it is well advised to consider the varying importance throughout the dataset during visualization and analysis. Importance-driven techniques can be classified into two categories: with and without direct user interaction. 

Studies that involve user interaction usually require users to decide the importance. Driven by the visualization goal, Peng et al.\cite{peng2013isosurface} proposed to define mesh importance using transfer functions for interactive isosurface rendering.
Burger et al.\cite{burger2008ImportParticle} proposed to control the shape and density of particles so that scientists can focus on the important regions where the region of interest is either user-defined or feature-based. 
Viola et al.\cite{Viola2006FocusAttention} defined the object of interest through user selection and smoothly modifies viewpoint and visual parameters when changing the object of interest.
Viola et al. \cite{viola2004imporVR} proposed importance-driven volume rendering by manually assigning different importance to the pre-segmented objects in the data to maximize the visual information in the rendered results. 
Wang et al. \cite{Wang2011FocusContext} proposed a feature-preserving data reduction method that allows users to magnify regions according to the degree of interest for focus+context visualization.

Importance-driven visualization without user interaction has pre-defined importance based on the domain knowledge or is totally data-driven.
Wang et al.\cite{wang2009AppDrCompTV} incorporated domain knowledge, e.g., salient isosurface and defined the importance of data based on the inverse distance to the surface of interest.   
To reduce massive visual information during particle tracing, Viola et al.\cite{Viola2005feature} utilized the object importance to define the sparseness level of each feature for controlling opacity values and rendering styles of the feature. 
Other works define data importance based on statistical models. For example, Wang et al.\cite{Wang2008ImportTV} defined importance through conditional entropy by measuring the amount of entropy one block remains given blocks of neighboring time steps.  
Gosink et al.\cite{gosink2010multivariateQDV} introduced a statistical framework to explore variable trends and identify important variables for different regions. 

Our work is related to importance-driven visualization and analysis. The difference is that we use the importance to generate a controlled latent representation.

\section{Background}
Our importance-driven latent generation framework is based on an autoencoder with a quantizer in the latent space. In this section, we introduce this model. 
\vspace{-1pt}
\subsection{Non-linear Transform Coding using Autoencoders} \label{NonLinearCodingAE}
A recent work~\cite{balle2020nonlinear} indicates that compared with linear transform coding, nonlinear transform coding is more flexible and can better adapt to the source signal distribution. In our work, we utilize nonlinear transform coding via a convolutional autoencoder. The autoencoder contains two parts, an encoder $f$ which converts the raw data $x$ into a latent representation $y$ and a decoder $g$ which decodes the latent $y$ and gets a reconstruction $\hat{x}$ of $x$. The latent size is often smaller than the raw data, which forms a bottleneck to
restrict the information flow from the encoder to the decoder. For example, as shown in \autoref{fig:entropyProb}, after several convolutional layers, the original data are converted to a latent of size $K\times 3\times 3\times3$, where $K$ is the number of filters in the last convolutional layer, also known as channel size of the latent. The bottleneck forces the latent to preserve only the most vital information in the data. Thus, the autoencoder is suitable to generate compact data representations. 
\vspace{-10pt}
\begin{figure}[htp]
    \centering
    \includegraphics[width=7cm]{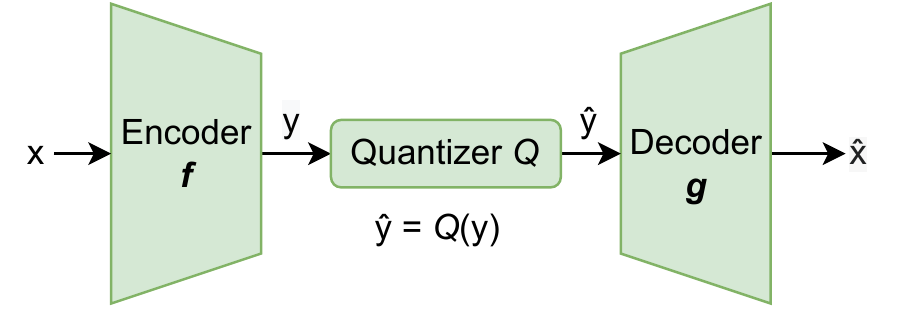}
    \vspace{-10pt}
    \caption{Autoencoder with a quantizer $Q$ in the latent space.}
    \label{fig:AEcomp}
\end{figure}
\vspace{-12pt}
\subsection{Quantized Latent Space for Data Reduction} \label{QuantizedLatentSpace}
To get a better data reduction performance, unlike a basic autoencoder which only minimizes the reconstruction loss, Ball{\'e} et al.~\cite{balle2016AEcomp} further quantize the latent representation by a quantizer $Q$, as shown in~\autoref{fig:AEcomp}. Then, the discrete symbols themselves are losslessly compressed through entropy coding. The optimization goal of this autoencoder is to minimize both the reconstruction loss and the entropy of the quantized latent representations, formulated as:
\vspace{-5pt}
\begin{equation} \label{eq:aecomp1}
\underbrace{\mathbb{E}_x[-\log _2 p_{\hat{y}}(\hat{y})]}_{R} + \lambda \underbrace{\mathbb{E}_x[\|x-\hat{x}\|^{2}_2]}_{D}
\end{equation}
where the quantized latent $\hat{y}=Q(f(x))$, the reconstructed data  $\hat{x}=g(\hat{y})$, and $Q$ is the quantizer. \autoref{eq:aecomp1} is aligned with the rate-distortion theory. $R$ is the rate that determines the number of bits per symbol for data reduction; in our case, it is the latent entropy. $D$ is the distortion between the original and the reconstructed data, i.e., reconstruction loss. $\lambda$ is a tradeoff parameter. A larger $\lambda$ will focus more on reducing the distortion $D$ during optimization. As a result, more bits are required to maintain the reconstruction quality, and we will have a larger rate $R$.

\begin{figure}[htp]
    \centering
    \includegraphics[width=8cm]{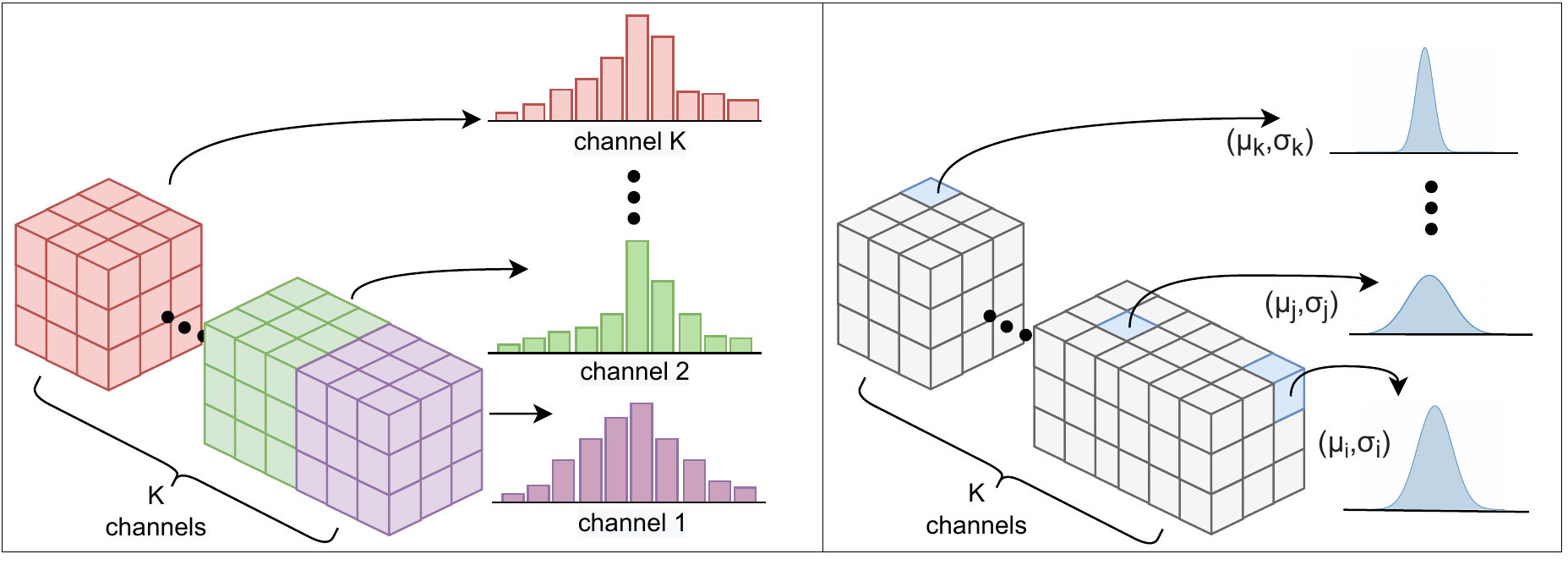}
    \vspace{-10pt}
    \caption{The probability estimation for every latent channel (left), and probability estimation for every latent dimension (right).}
    \label{fig:entropyProb}
\end{figure}

However, the quantization is not differentiable.
To make quantization differentiable and incorporate the quantization error during training, Ball{\'e} et al.~\cite{balle2016noise} replace the quantizer with additive uniform noise. Now instead of the quantized representation $\hat{y}=Q(f(x))$, we have a ``noisy'' representation $\tilde{y}=f(x)+\Delta y$, where $\Delta y\sim U(-\frac{1}{2}, \frac{1}{2})$. The optimization goal changes into~\cite{balle2016AEcomp}:
\vspace{-4pt}
\begin{equation} \label{eq:aecomp2}
\underbrace{\mathbb{E}_{x,\Delta y}[-\log_2 p_{\tilde{y}}(\tilde{y})]}_{R} + \lambda \underbrace{ \mathbb{E}_{x,\Delta y}[\|x-\tilde{x}\|^{2}_2]}_{D}
\end{equation}
\vspace{-7pt}

\noindent where $\tilde{x}=g(\tilde{y})$ is the reconstruction. To remove the constraint on the input size, a non-parametric distribution is used to model the probability density for channels of latent $p_{\tilde{y}}$, as shown in~\autoref{fig:entropyProb} (left). 
During testing, the actual quantization, such as rounding is applied. After that, a lossless entropy encoding on the quantized latents is applied to convert latents into bitstreams. More frequent data will be represented by shorter bits than less frequent data. 
\vspace{-6pt}
\vspace{-5pt}

One limitation of the above method is that the entropy estimation of the latent representation is not accurate. Entropy encoding relies on the probabilistic distributions of discrete latents to decide which codeword will represent which quantized symbol so that the average bit length is minimal. The better the probabilities are modeled, the closer the bit rate approaches the optimal lower bound. 
However, the above method does not consider spatial and raw data dependencies when estimating the probability~\cite{balle2018hyperprior}, due to the reason that it only models a channel-wise latent distribution for an ensemble of input. 

To improve entropy estimation, one follow-up work \cite{balle2018hyperprior} introduces a hyperprior network to extract side information to assist latent probability estimation. The hyperprior network takes side information as input to predict a prior on the parameters of latent's probability distribution. As shown in~\autoref{fig:entropyProb} (right), each latent dimension is modeled as a Gaussian where the scale and mean of each Gaussian are predicted by the side information~\cite{balle2018hyperprior}.

With the quantized latent space and improved entropy estimation, we can achieve a better data reduction performance. In the next section, we will present our latent generation method based on a quantized autoencoder which achieves the data reduction goal, and more importantly, takes domain interest into consideration during latent generation. 


\section{Overview and Algorithm Requirements}
\textbf{Algorithm Requirements:}
We summarize three algorithm requirements to generate domain-interest-guided latent representations.
\begin{itemize}
    \item The generated latent representations need to respond to different domain interests.  
    \vspace{-5pt}
    \item The algorithm needs to be adaptive to different types of domain interests such that scientists do not need to train multiple neural network models when they vary their interests. 
    \vspace{-5pt}
    \item The algorithm needs to generate compact latent representations whose size depends on the domain interest, i.e., low domain interest means a more compact latent representation. 
\end{itemize}

\begin{figure}[htp]
    \centering
    \includegraphics[width=7.5cm]{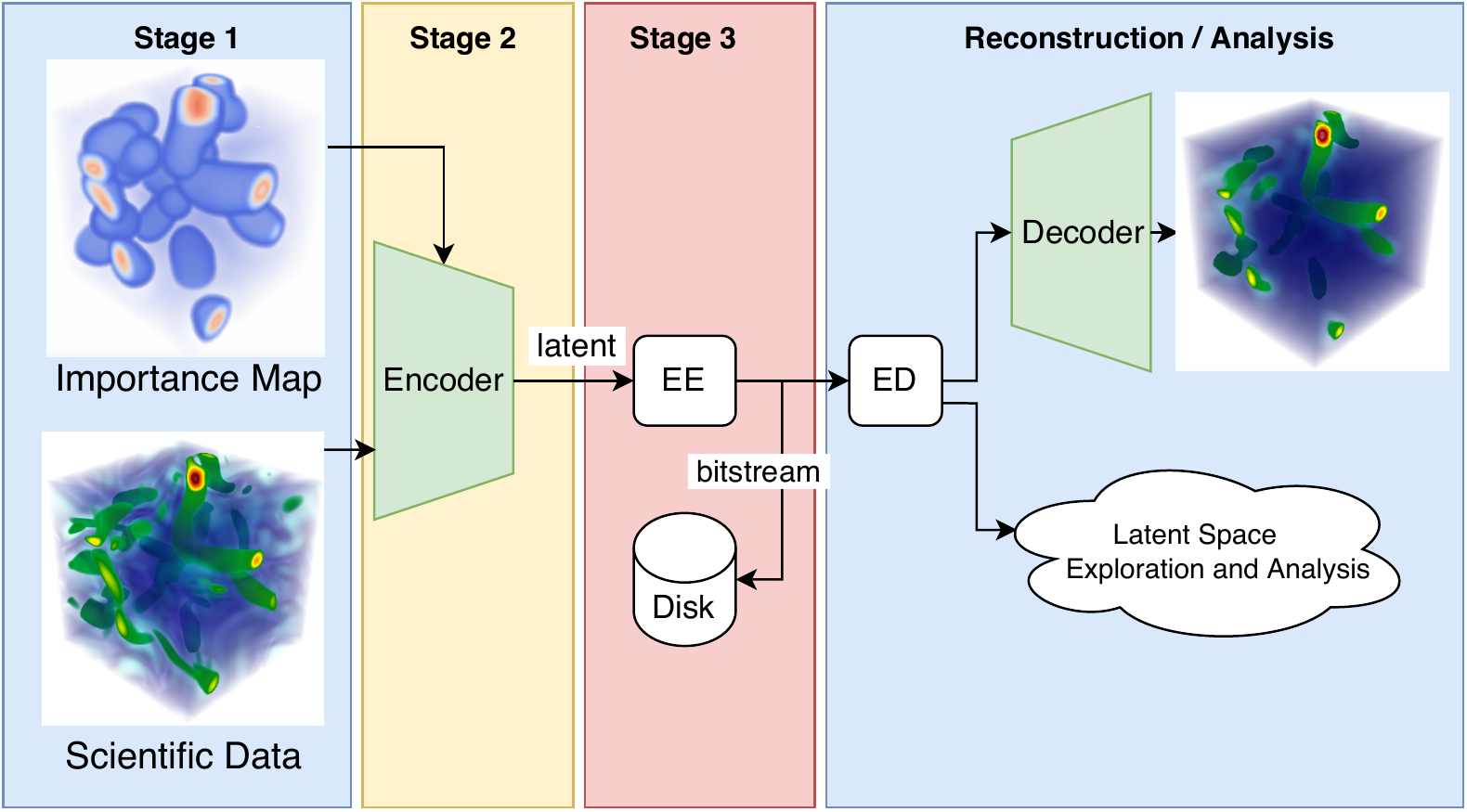}
    \vspace{-8pt}
    \caption{Our proposed method to generate importance-driven latent representations conditioned on the input importance maps. The resulting latent representation will be entropy encoded into bitstreams for saving. }
    \label{fig:framework}
\end{figure}

\textbf{Overview}
To generate latent representations for scientific data guided by scientists' interests, we propose an importance-driven latent generation algorithm. An overview of the proposed method is shown in \autoref{fig:framework}.
The first stage of our method is to properly represent various scientific interests with spatial importance maps, which can be interpreted as, for each spatial location, how much information scientists want to preserve when generating the compact latent representation. 
Then the second stage is to generate scientific interest-guided latent representations through our autoencoder model. We take a block-wise processing strategy. Volumetric data and corresponding importance maps are divided into blocks and then processed by the model. Conditioned on the importance map, the data blocks are non-linearly encoded and transformed by the autoencoder’s encoder into compact latent representations.
The third stage is a lossless data reduction component in the latent space. Inspired by autoencoders used for image compression \cite{balle2018hyperprior, balle2016AEcomp} as discussed in \autoref{QuantizedLatentSpace}, our latent representations are further quantized into discrete symbols. After that, an entropy encoding algorithm, e.g., Asymmetric Numeral Systems (ANS), is adopted to losslessly compress quantized latents into bitstreams for saving.

Analyses can be done in either latent space or the original physical space. Importance-driven latent space has a simpler structure, and therefore tasks like feature extraction can be easily performed in this space. When the precise visualization of the dataset is needed, latent representations can be decoded back into the physical space for various visualization tasks. 
\vspace{-6pt}
\vspace{-12pt}

\section{Method}
There are three main issues to address when generating domain-interest-guided latent representations: (1) how to define a unified representation to incorporate various domain knowledge and avoid training multiple models when scientists vary their interests, (2) how to fuse domain knowledge during latent generation, and (3) how to control the latent size and make trade-offs between latent size and latent quality. In this section, we discuss how we address these issues. 

\subsection{Importance Driven Latent Generation}
\subsubsection{Domain Interest Representation}\label{sect:Representation}
This section discusses how to represent diverse domain interests with a unified representation. Our latent generation method is domain-interest-guided, meaning not all data are treated equally important during latent generation. This strategy is commonly used for  many scientific data analysis applications. 
Given a massive amount of scientific data, scientists often try to identify features of interest by narrowing down their search space, which is defined by how data are relevant to the important features. Generating latent representations for scientific data according to the data importance can not only reduce the size of data, but also allow scientists to focus on the most salient portion of the data.  


In our method, we utilize data importance to assist the process of latent generation. To create a unified representation of importance that can be taken by our autoencoder for a variety of needs, we summarize commonly used importance definitions in scientific visualization literature as below:
\begin{itemize}
  \item \textbf{Location-based:} Scientists assign an importance value for every spatial location based on whether it is in the pre-selected region of interest~\cite{Viola2006FocusAttention, Viola2005feature, viola2004imporVR}. For example, hurricane eye regions are of high interest for hurricane and tropical cyclone research scientists, and as a result, those regions will have higher importance values than non-hurricane eye regions~\cite{Viola2005feature}.
  
  \vspace{-6pt}
  \item \textbf{Distance-based:} To enhance the understanding of features of interest, scientists define the importance of each data element based on its distance to the feature of interest. For example, based on the distance to object of interest~\cite{Viola2005feature} or salient isosurfaces~\cite{wang2009AppDrCompTV}.
  
  \vspace{-6pt}
  \item \textbf{Value-based:} Scientists define the importance based on the differences between a pre-selected reference value and data values or based on transfer functions~\cite{Wang2011FocusContext, peng2013isosurface}. If the value of a data element is close to the values of interest, high importance will be assigned. We note that here the value can take a variety of forms: scalars, vectors, and tensors, to name a few. 
  
  \vspace{-6pt}
  \item \textbf{Time-based:} To effectively visualize and analyze large-scale time-varying data, scientists assign different importance values for time steps. For example, based on the relative information a time step contains about its temporal neighbours~\cite{Wang2008ImportTV} or assign salient time steps with high importance values.
  
  \vspace{-6pt}
  \item \textbf{Multivariate-based:} For a multivariate dataset, the importance can be derived from the joint distribution of variables~\cite{gosink2010multivariateQDV}. For instance, to locate interesting regions for turbulent combustion data, multiple variables such as Mixture Fraction (MIX), Mass Fraction of the Hydroxyl Radical (OH), and Heat-Release Rate (HR) are jointly considered. 
  
  \vspace{-6pt}
  \item \textbf{Statistical-based:} Importance can be defined based on statistical properties of data such as conditional entropy~\cite{Wang2008ImportTV}, correlation, or value histogram~\cite{biswas2020probabilistic}. 
\end{itemize} 
\vspace{-3pt}

In the core of our method, we define a unified representation, i.e., a real-valued spatial importance map, to incorporate various importance definitions. The map is defined in the same domain as the data, and each value in the importance map indicates the scientific interest at that spatial location. Importance maps are taken as an additional input to the neural network model to control the latent representation generation.

There are two obvious advantages in using spatial importance maps. First, spatial importance maps can inform the neural network which regions are more important so that their information needs to be better preserved in the latent space, and for those regions with low importance, their latents can be simplified or smoothed out during encoding. Second, because we are using a unified representation for various domain interests and our latent generation is conditioned on the input importance map, as a result, we do not need to retrain different neural network models when scientists change their definition of spatial importance. 
In our paper, the importance value at every spatial location is calculated through a scientist-specified importance mapping function $\Psi$, mathematically defined as:
\vspace{-4pt}
\begin{equation} \label{eq:Importance}
I_p := \Psi(p,F(p))
\end{equation}
\vspace{-12pt}

\noindent where $\Psi\colon \mathbb{R}^3 \to \mathbb{R}$ is a mapping from spatial location $p\in \mathbb{R}^3$ to an importance value $I_p$ given the location and its data value $F(p)$.  $F(p)\in \mathbb{R}$ if it is a univariate data, and $F(p)\in \mathbb{R}^n$ if it is a multivariate data with $n$ variables. We evaluate $I_p$ on all voxel locations to obtain an importance map $I$. 

During training, we randomly generate importance maps with different spatial variations such as distance ramps, Gaussian distributions with various centers, data gradients, and random uniform maps. During testing, the trained model is applied to various importance maps derived from different scientific interests. In our evaluation in \autoref{sect:evaluation}, we demonstrate that these predefined importance maps are effective to train a generalized model which does not constrain a scientist's importance map specification during testing. 

\subsubsection{Autoencoder with Condition Network}
To generate reduced data representations, we utilize a convolutional autoencoder model which converts input data into a latent representation through an encoder and decodes the latent back to data through a decoder. We utilize autoencoder for the reason that, as also discussed in \autoref{NonLinearCodingAE}, compared with linear coding methods such as discrete cosine transform (DCT), the non-linear coding ability of autoencoders makes them suitable and powerful to represent data. 

To properly fuse domain knowledge into latent representations, 
we utilize Spatial Feature Transform (SFT) layers which are widely used in computer vision for image super-resolution~\cite{wang2018SFT, gu2019blindSR, Chen2021SpatialAttention}, conditional generation~\cite{wang2021FaceRestoration, wang2019edvr}, compression~\cite{song2021variable}, and segmentation~\cite{Li2021Segmenting}. In these works, SFT layers are used to incorporate conditional knowledge by generating affine transformation parameters for feature modulation. We adopt a similar method as Song et al.~\cite{song2021variable} which performs image compression given a classification or text-preserving task. 

\begin{figure}[htp]
    \centering
    \includegraphics[width=6.5cm]{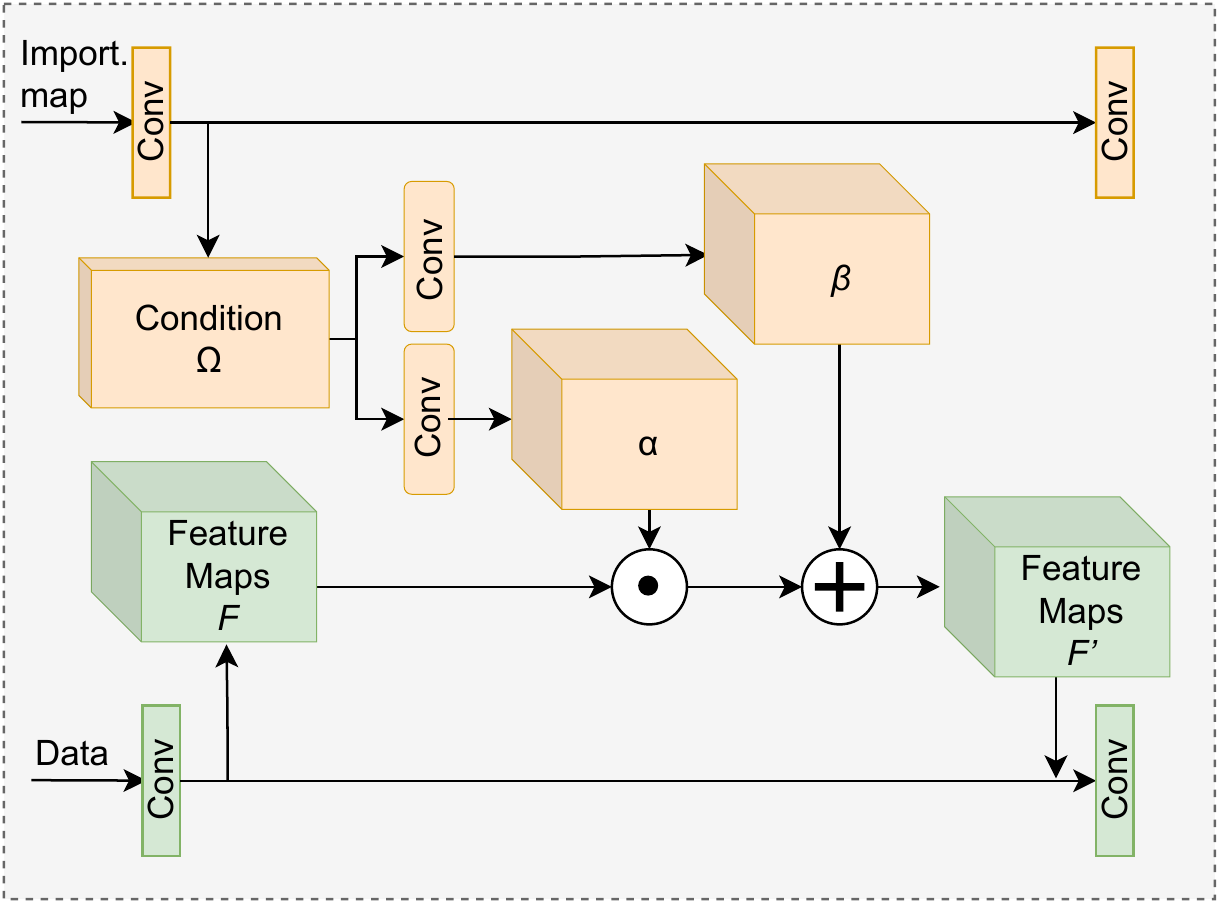}
    \vspace{-9pt}
    \caption{Spatial Feature Transform (SFT)~\cite{wang2018SFT} layers take the condition $\Omega$ generated from the importance map as input and output affine transformation parameters to scale ($\alpha$) and shift ($\beta$) feature map $F$ of data.
    }
    \label{fig:sft}
\vspace{-5pt}
\end{figure}

In our work, we utilize SFT layers to fuse domain knowledge into latent representations. The reason for using SFT layers is that they can capture rich spatial prior information from prior knowledge, e.g., regions of interest, to modify the intermediate feature maps of the data in the autoencoder. For example, smoothing out details in regions where scientists have low interest.
More specifically, we adopt the autoencoder model by connecting it with two SFT-layer-based feature transformation networks. 
The transformation network connected to the encoder is shown in \autoref{fig:sft} who takes domain interest (i.e., a spatial importance map $I$) as input to extract conditions of different resolutions to have a layer-by-layer control of the encoder. Each condition $\Omega$ produces two sets of affine transformation parameters (i.e., $\alpha$ for scaling and $\beta$ for shifting) for each encoder layer, formulated as:
\vspace{-4pt}
\begin{equation}
\Omega = conv(I)
\end{equation}
\vspace{-14pt}
\begin{equation}
\Phi(\Omega) = (\alpha, \beta)
\end{equation}
\vspace{-11pt}

\noindent where $conv$ are the convolutional layers. $\Phi$ is a mapping function from condition $\Omega$ to the scaling parameter $\alpha$ and the shifting parameter $\beta$.
$\alpha$ and $\beta$ are used to transform the intermediate feature map $F$ generated by the autoencoder:
\begin{equation} \label{eq:SFT}
F'=\textnormal{SFT}(F|\alpha, \beta) = F\odot\alpha+\beta
\end{equation}
\vspace{-12pt}

Then, the transformed feature map $F'$ is taken as input to the next encoding layer. $\alpha$ and $\beta$ are of the same size as the feature map $F$, and $\odot$ denotes element-wise multiplication. 
Applying multiplication and addition on feature maps is a simple and effective way to gradually fuse two sources of information (i.e., importance and data) from different levels. Scaling the feature map is like gating so that information in regions with high importance are preserved and others are suppressed. Shifting the feature map has a similar effect.
Combining these two, we have the flexibility to leverage importance maps for latent generation.

\vspace{-8pt}

\subsection{Entropy Encoding in Latent Space}
To optimize the size of the latent representation, we apply lossless entropy encoding to the latent vectors generated from the autoencoder. We adopt the entropy model widely used for neural-network-based image compression \cite{balle2018hyperprior}, as discussed in~\autoref{QuantizedLatentSpace}. 

\begin{figure}[htp]
    \centering
    \includegraphics[width=9cm]{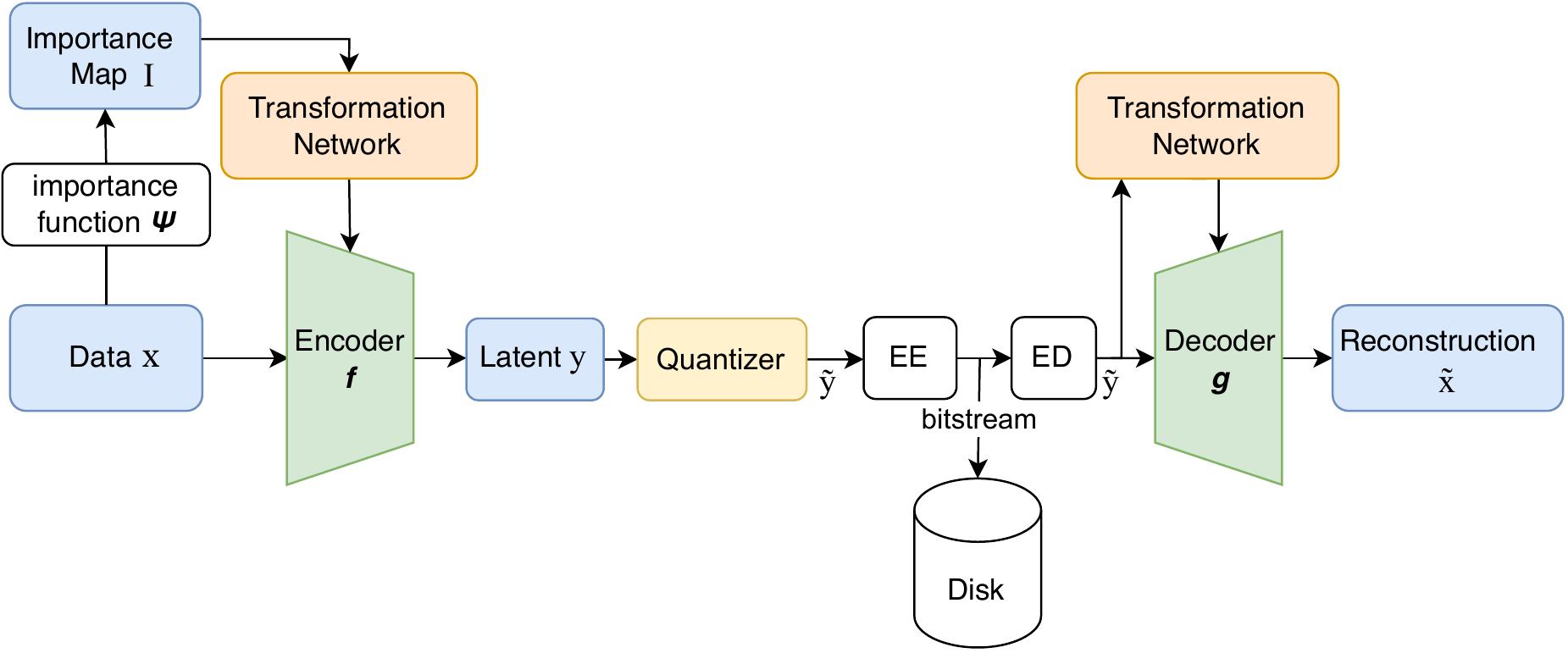}
    \vspace{-18pt}
    \caption{Our proposed importance-driven latent generation method. Our model is a combination of an autoencoder and entropy encoding model. }
    \label{fig:overview}
\end{figure}

Our proposed importance-driven latent generation method is shown in~\autoref{fig:overview}. After layers of non-linear coding using autoencoder's encoder $f$, the input data $x$ is converted into a latent representation $y$, controlled by a scientist-specified input importance map $I$, so we have:

\vspace{-5pt}
\begin{equation}
y = f(x, T_f(I))
\end{equation}
\vspace{-12pt}

\noindent where $T_f$ is the transformation network connected to the encoder $f$. Following the technique in \autoref{QuantizedLatentSpace}, we quantize the latent vector $y$ and apply the entropy encoding algorithm, Asymmetric Numeral Systems (ANS)~\cite{Duda2015ANS}, on latent vectors. The resulting bitstreams are saved into the disk. During decoding, the saved bitstreams are entropy decoded into discrete latent $\tilde{y}$ and sent to the decoder to get the reconstruction $\tilde{x}$. As shown in~\autoref{fig:overview}, to save storage and mainly reduce unimportant information in the latent representation, we do not use importance maps to modulate information during decoding. The decoder only takes latent $\tilde{y}$ as input to the transformation network connected to the decoder to incorporate conditions during reconstruction, formulated as:  
\vspace{-4pt}
\begin{equation}
\tilde{x} = g(\tilde{y}, T_g(\tilde{y}))
\end{equation}
\vspace{-12pt}

\noindent where $T_g$ is the transformation network connected to the decoder $g$. 

\vspace{-6pt}
\vspace{-12pt}

\subsection{Loss Functions} \label{sect:lossFunction}
To make trade-offs between latent size and quality based on domain interests, we formulate our importance-driven latent generation as an optimization problem of minimizing the rate-distortion Lagrangian as discussed in \autoref{QuantizedLatentSpace}, formulated as: 

\vspace{-4pt}
\begin{equation} \label{eq:RDloss}
    \mathcal{L} = \mathcal{L}_{R} + \lambda\mathcal{L}_{D}
\end{equation}
\vspace{-10pt}

\noindent where $\mathcal{L}_{R}$ is the quantized latent entropy loss, $\mathcal{L}_{D}$ is the reconstruction loss, and $\lambda$ is the Lagrangian multiplier, a hyperparameter to balance between reconstruction quality and latent size.

To force the generated latent representation to be guided by scientists' spatial interests, instead of a reconstruction loss with equal importance on every spatial location like a basic autoencoder, we use an adaptive reconstruction loss between the input data $x$ and reconstructed data $\tilde{x}$:
\vspace{-5pt}
\begin{equation} \label{eq:Dloss}
    \mathcal{L}_{D} = \mathbb{E}_{x,\Delta y}
[\sum_{i=1}^{N}w_i(x_i-\tilde{x}_i)^2]
\end{equation}
\vspace{-9pt}

\noindent where $w_i$ is the distortion weight at index $i$ derived from the importance map $I$ by an exponential function with a hyperparameter $a$, i.e., $w_i=e^{aI_i}$. The goal of exponential weighting is to have a finer distortion control across different spatial locations such that different importance values will have significantly different contributions to $\mathcal{L}_{D}$. $N$ is the number of data elements in $x$. $\Delta y$ is the additive uniform noise to relax quantization. 

The entropy loss for discrete latent $\tilde{y}$ is:
\vspace{-5pt}
\begin{equation} \label{eq:Rloss}
    \mathcal{L}_{R} = \mathbb{E}_{x,\Delta y}[-\log_2 p_{\tilde{y}}(\tilde{y})] 
\end{equation}
\vspace{-12pt}

\noindent where $p_{\tilde{y}}$ is the latent probability distribution for entropy coding. 

\subsection{Latent Space Analysis}
\label{sect:LSA}
In this section, we discuss how we use block-wise latent representation for feature-related analysis and briefly introduce the visual exploration tool for latent spaces. 
Feature-related analysis in the latent space originated from the observation that similarity defined in the latent space can better represent the similarity of higher-level features than the similarity of the raw data \cite{bengio2013representationLearn}. 
Our latent vector generation method considers domain knowledge provided as an importance map to neural networks. As a result, latent space distribution is conditioned on the provided importance map. Even though the structure of the full latent space of the dataset is complicated and requires a sophisticated tool \cite{li2022local} to explore, importance-driven latent space is easier to understand and to be made into use for feature analysis.

On the saved latent representations for each data block, we perform a hierarchical clustering algorithm to identify blocks that are similar to each other. These clusters can be either used to extract a subset of data for further analysis or for feature-driven visualization like the one presented by Cheng et al. \cite{cheng2018deep}. It is worth noting that the data block size is a hyperparameter we should choose based on the feature size, data complexity, and desired representation storage size.

Our latent space visual analysis tool is based on the one designed by Li and Shen \cite{li2022local}, where clustering is performed on block-wise latent vectors and cluster results are visualized in latent, and physical space for feature-related analysis. 
There are three main views in our tool, \textit{hierarchical clustering view}, \textit{latent space view}, and \textit{physical space view}, demonstrated in \autoref{fig:iso8}. \textit{Hierarchical clustering view} presents each cluster as a node in a tree graph, where clusters can be modified by interacting with the nodes. \textit{Latent space view} shows projected latent vectors into 2D using t-distributed stochastic neighbor embedding (t-SNE) projection~\cite{Laurens:2008:tsne}. This view is updated when the clustering result is modified. Finally, \textit{physical space view} visualizes data in the selected cluster.
Spectral clustering is used in our latent analysis approach since it adapts well to complex spaces with unknown cluster shapes \cite{shi2000normalized}, which is usually the case for latent spaces generated by neural networks. 





\subsection{Implementation}
\subsubsection{Block-based Processing} 
\label{sect:block_size}
Processing large-scale scientific data requires a big convolutional neural network model which has high computational cost and memory consumption. Another problem is that we need a large collection of data for training, but building such training data is prohibitive for scientific simulations due to the high cost of generating and saving large-scale data. 
To meet GPU memory constraints, some prepossessing steps such as downsampling or cropping~\cite{Han2020FlowNet, Han2021V2V} are applied to data. 
However, the drawback of downsampling is that it inevitably introduces errors and uncertainties in the downsampled data. 

To address the above problems, we adopt a block-based processing strategy, i.e., volumetric data are divided into blocks for the neural network model to process. Data blocks can be processed in parallel with a large batch size for speedup. To reduce the reconstruction error at the block boundary introduced by zero-padding or reflection-padding, we pad each block with the actual data for the network to process. During reconstruction, we crop the reconstructed data, and only the central data regions are attached to reconstruct the whole volume. For instance, if the block size is $24^3$ and we have a padding size equal to $4$, then the data size each latent represents is $16^3$. 

\subsubsection{Training Data Sampling}
The most intuitive way to build the training data is to randomly sample large amounts of data blocks to ensure a good coverage of different patterns for training. However, this data size will grow proportionally to the size of the original data and make the training extremely ineffective. To solve this and to force the model to learn complex patterns, we adopt a complexity-aware training data sampling strategy, i.e., the training dataset is designed to include more complex (high entropy) blocks and less homogeneous (low entropy) blocks. For the Hurricane Isabel dataset, the sampling ratio between high and low entropy blocks is 10:1 to ensure the complex data regions are covered in the training data given that a large portion of the original data is homogeneous.

\begin{table*}[!ht]
\small
\caption{\label{tab:Datasets}Dataset name, variable name, data resolution, training epochs and time, training data size (number of blocks), encoding and decoding time on each volume (seconds). Encoding and decoding time do not induce the time used for writing/reading bitstreams files into/from disk.
 }
\vspace{-2pt}
\centering
 \begin{tabular}{c|c|c|c|c|c|c|c} 
 Dataset & Variable & Size & Epochs & Training time & \# Training blocks & Enc. time & Dec. time \\ [0.5ex] 
 \hline
 Vortex & vorticity magnitude & 128$\times$128$\times$128 & 600 & 4h 20m & 1000 & 0.0232s & 0.0105s \\
 Nyx & log density & 256$\times$256$\times$256 & 100 & 3h 14m & 5000 & 0.2272s & 0.0794s\\
 Isabel & pressure & 512$\times$512$\times$96 & 200 &  5h 30m & 5500 & 0.2000s & 0.1191s\\[0.5ex]
 \end{tabular} 
 \label{table:dataset}
\vspace{-9pt}
\end{table*}

\section{Results} \label{sect:evaluation}
In this section,  we evaluate the  \textbf{I}mportance-\textbf{D}riven \textbf{Lat}ent generation method (\textbf{\sysname}) 
both quantitatively and qualitatively
from four different perspectives: (1) the quality of latent representations;  (2) the size of latent representations; (3) the influence of different important maps; and (4) the use of latent representations for latent space analysis.  

\subsection{Dataset and Training Parameters}
We evaluated our importance-driven latent generation method using three scientific datasets for multiple scientific applications.

\textbf{Vortex} is a simulation of vortex structures with spatial resolution 128$\times$128$\times$128 across 30 time steps. We used the vorticity magnitude scalar field for experiments. We randomly sampled 1000 data blocks from 5 time steps as the training data. 
\textbf{Nyx} is a cosmological simulation produced by Lawrence Berkeley National Laboratory. We used the log density field with resolution 256$\times$256$\times$256. 5000 data blocks from 5 ensemble members were randomly sampled for training. 
\textbf{Hurricane Isabel} is a simulation of Hurricane Isabel, produced by the Weather Research and Forecast (WRF) model, courtesy of NCAR and the U.S. National Science Foundation (NSF). The data were sliced along the z dimension to remove the special value $1 \times 10^{35}$ representing "no data" (the land region). In our experiment, the resolution of data is 512$\times$512$\times$96 with 48 time steps. We chose the pressure field for evaluation. Training data contain 5500 data blocks from 5 time steps. 

Our work consists of two main components: the {\sysname} model and a latent space visual analysis tool. 
The {\sysname} model is implemented based on PyTorch$\footnote{https://pytorch.org}$ and trained with a single NVIDIA Tesla P100 GPU. For all datasets, we use Adam optimizer~\cite{Diederik2015Adam}. The learning rate for the autoencoder model and the entropy model is $10^{-4}$ and $10^{-3}$, respectively. Total training time for each dataset is listed in ~\autoref{table:dataset}.
Based on a fully convolutional model with block-based training and inference strategy, we can apply {\sysname} on data of any resolution. 
The hierarchical clustering view and latent space projection view from the latent space exploration tool are implemented with Vue.js$\footnote{https://vuejs.org}$ as the front-end framework and Flask$\footnote{https://flask.palletsprojects.com}$ as the back-end framework. VTK APIs$\footnote{https://vtk.org}$ are used to visualize the extracted blocks.

\subsection{Quantitative Evaluation} \label{sect:QuantiEval}
In this section, we quantitatively evaluate the size and the quality of latent representations generated by {\sysname}. 
\subsubsection{Evaluation Metric}
To evaluate the size of latents, we use the ratio between the original data size and the saved bitstream file size, i.e., latent size ratio (LSR):
\vspace{-3pt}
\begin{equation} \label{eq:lsr}
LSR=\frac{\text{original data size}}{\text{bitstream file size}}
\end{equation}
\vspace{-7pt}

To evaluate the quality of importance-driven latent representations, we analyze how well the important regions are preserved during reconstruction under various importance maps. We compute the error between the decoder's reconstruction and the raw data. Because we focus more on the quality of important regions, we utilize a weighted Mean Squared Error (wMSE) defined as:
\vspace{-4pt}
\begin{equation} \label{eq:wmse}
\textnormal{wMSE}(x, \tilde{x}) = \frac{1}{\sum_{i=1}^{N}I_i} \sum_{i=1}^{N}I_i(x_i-\tilde{x}_i)^2
\end{equation}
\vspace{-7pt}

\noindent where $x_i$, $\tilde{x}_i$ are the original and the reconstructed data at position $i$, respectively. $N$ is the number of data elements in $x$. $I_i$ is the importance at position $i$ in range [0, 1] defined by~\autoref{eq:Importance} for different applications. Locations with a larger importance value will have higher weights in the error estimate. The peak signal-to-noise ratio (PSNR) is defined based on wMSE:
\vspace{-4pt}
\begin{equation} \label{eq:wpsnr}
\textnormal{PSNR}(x, \tilde{x}) = 10\log_{10} \frac {v^2}{\textnormal{wMSE}(x, \tilde{x})} 
\end{equation}
\vspace{-9pt}

\noindent where $v$ denotes the value range in the original data. 


\subsubsection{Evaluation on Different Importance Maps} \label{sect:map_defination}
As discussed in~\autoref{sect:Representation}, there are various criteria focusing on representing regions of interest for importance-driven visualization. In this section, we evaluate {\sysname}’s quality and quantity under different importance definitions. 


To evaluate the effectiveness of the entropy encoding module in {\sysname}, especially its ability to generate the latent representation of optimal size with the presence of an importance map, we compare {\sysname} with a baseline method, i.e., a basic autoencoder without the importance map, quantization, entropy module, and the entropy loss. We train this baseline model with the same training data and parameter setting as {\sysname}, but only with the reconstruction loss. 

\textbf{Distance-based importance maps:} We evaluate importance-driven latent's quality and quantity conditioned on distance-based importance maps through the Vortex dataset. Vortex data contains vortex structures that have been widely used for isosurface tracking to analyze vortex core regions over time~\cite{ji2006FTEMD, Ji2003HighDimTrack}. In this evaluation, the importance maps are defined based on distances to the selected isosurfaces where the importance value $I_p$ for each spatial location $p$ defined in~\autoref{eq:Importance} can be specified as: 
\vspace{-4pt}
\begin{equation} \label{eq:VortexImp}
I_p := \Psi _{Vor}(p) =  e^{-0.2|SDF(p,S)|}
\end{equation}
\vspace{-10pt}

\noindent $SDF(p,S)$ represents the signed distance from location $p$ to surface $S$. We use a negative exponential function to convert absolute SDF distances into importance values in [0, 1] for the model to process. The importance will decrease exponentially as $p$ becomes far from the surface. $0.2$ is a parameter that we choose to have a proper slope of the exponential curve. We can also use other functions to convert distances into weights, e.g., the inverse distance function.

To evaluate the influence of different distance-based important maps, we chose a time step in Vortex data and pre-selected several vorticity magnitude values (e.g., 5.0, 6.0, 7.0, and 8.0) as salient isovalues to be preserved in the latent representations. We used $\Psi _{Vor}$ in~\autoref{eq:VortexImp} to convert the volumetric data into importance maps for {\sysname} to generate importance-driven latent representations. 

We report the quantitative results, i.e., PSNR and latent size ratio (LSR), of applying different distance-based importance maps on Vortex data in~\autoref{tab:eval_vortex_nyx}. From the table, we found that {\sysname} indeed can generate importance-driven latent representations for Vortex. Given the same data but different importance maps, {\sysname} generates latents with different quality and size, as shown in the PSNR ({\sysname}) and LSR ({\sysname}) columns. As for the baseline model, because the latent size is determined by the input data size, the latent size ratio for the baseline model is fixed at a very small number ($7.3143$) as shown in LSR (Base). 
It is clear in the table that the proposed {\sysname} can largely decrease the latent size (i.e., increase LSR) based on different importance maps without losing much quality.  
We also found that {\sysname} can achieve increased latent size ratios as we increase the target isovalue. This is because {\sysname} is designed to have higher quality on regions with high interests, and as the isovalue increase, the important regions are getting smaller for this dataset, and as a result, LSR is getting larger. 

\textbf{Value-based importance maps:} Given value-based importance maps, we evaluate the size and quality of latents generated by {\sysname} on Nyx data. One important post-hoc analysis task for Nyx simulation is to find dark matter halos which are related to the high-density field in the data~\cite{friesen2016situ}. In our experiment, the importance value $I_p$ of each spatial location $p$ is defined based on the log density value $F(p)$ and a reference log density value $F_{ref}$ as follows:
\vspace{-6pt}
\begin{equation} \label{eq:NyxImp}
I_p := \Psi _{Nyx}(p, F(p)) = 
\begin{cases}
1 & \text{if } F(p)>F_{ref}\\
0 & \text{else} 
\end{cases}
\end{equation}
\vspace{-10pt}

To evaluate the effect of different value-based importance maps, driven by the domain interests of the Nyx data discussed above, we conduct experiments on two ensemble members of Nyx with different reference values. In our experiment, we select $9.9$, $10.2$, and $10.5$ as log density reference values and utilize $\Psi _{Nyx}$ defined in~\autoref{eq:NyxImp} to compute value-based importance maps. 

We report the quantitative results in the second and the third blocks of rows of \autoref{tab:eval_vortex_nyx}. The two blocks of rows represent experiments on two different ensemble members, denoted as Nyx (m1) and Nyx (m2). As we can see, for each ensemble member, the latent size ratio (LSR ({\sysname})) increases as we increase the log density reference value (i.e., reduce the number of important voxels), but the quality (PSNR ({\sysname})) is kept comparable to the baseline (PSNR (Base)). Another thing we notice is that PSNR drops as we increase the log density reference value for both baseline and {\sysname} models. A possible reason can be that these high-value regions are harder to model due to high data complexity.

\textbf{Location-based importance maps:} 
We also evaluate the latent representations' size and quality given location-based importance maps using the Isabel dataset. For this dataset, one task that scientists are interested in is to identify and analyze the hurricane eye region. So the importance maps are built based on the hurricane eye locations where voxels inside the interested region $C$ will have high importance values. The importance value for each spatial location $p$ is defined as:
\vspace{-7pt}
\begin{equation} \label{eq:IsabelImp}
I_p := \Psi _{Isa}(p) = 
\begin{cases}
1 & \text{if } p\in C\\
0 & \text{else} 
\end{cases}
\end{equation}
\vspace{-10pt}

We use the first time step of Isabel and draw a bounding box of the hurricane eye as the region of interest $C$. Based on the importance mapping function in \autoref{eq:IsabelImp}, we compute the importance map.

\autoref{tab:eval_vortex_nyx} shows the quantitative results of Isabel data given a location-based importance map using {\sysname} and the baseline model. Compared with the baseline, the latent representation generated by {\sysname} with a location-based importance map is more compact (higher LSR) with slightly higher quality (higher PSNR) than the baseline. 


Essentially, the baseline model is a special case of our {\sysname} with $\lambda\rightarrow\infty$ in~\autoref{eq:RDloss}, which achieves an upper bound for reconstruction error and a lower bound for the latent size ratio. 
Instead of a static model with a fixed latent size, the proposed {\sysname} can achieve various quality and latent vector sizes based on target scientific applications.

\begin{table}[!t]
\small
\caption{ Reconstruction PSNR and latent size ratio (LSR) for {\sysname} with different importance maps and for the baseline model.}
\centering
 \begin{tabular}{p{8mm}|p{15mm}|*{3}{p{8mm}|}p{8mm}}
 Data & Imp. Def. & PSNR ({\sysname}) & PSNR (Base) & LSR ({\sysname}) & LSR (Base) \\ [0.5ex] 
  \hline
  \multirow{4}*{Vortex} & iso 5.0 & 34.3987 & 35.4927 & 107.3825 & \multirow{11}*{7.3143} \\ 
  ~ & iso 6.0 & 33.9121 & 35.2154 & 109.8901 & ~\\ 
  ~ & iso 7.0 & 33.2551 & 34.5266 & 111.849 & ~ \\ 
  ~ & iso 8.0 & 33.9757 & 34.2463 & 113.6767 & ~ \\
  \cline{1-5}
  
  \multirow{3}*{Nyx (m1)} & log den $>9.9$ & 33.8422 & 35.2382 & 209.4241 & ~\\ 
  ~ & log den $>10.2$ & 30.9899 & 31.0386 & 218.8783 & ~ \\ 
  ~ & log den $>10.5$ & 28.2268 & 27.1545 & 225.6700 & ~\\ 
  \cline{1-5}
  
  \multirow{3}*{Nyx (m2)} & log den $>9.9$ & 31.3774 & 32.2409 & 172.5998 & ~  \\
  ~ & log den $>10.2$ & 29.0695 & 29.2020 & 179.3722  & ~  \\
  ~ & log den $>10.5$ & 26.761 & 26.1568 & 188.2353  & ~ \\
  \cline{1-5}

  Isabel & hurricane eye & 44.9749 & 44.6605 & 199.1288 & ~  \\ [1ex] 
 \end{tabular}
 \label{tab:eval_vortex_nyx}
\vspace{-16pt}
\end{table}

\begin{figure}[htp]
    \centering
    \includegraphics[width=\columnwidth]{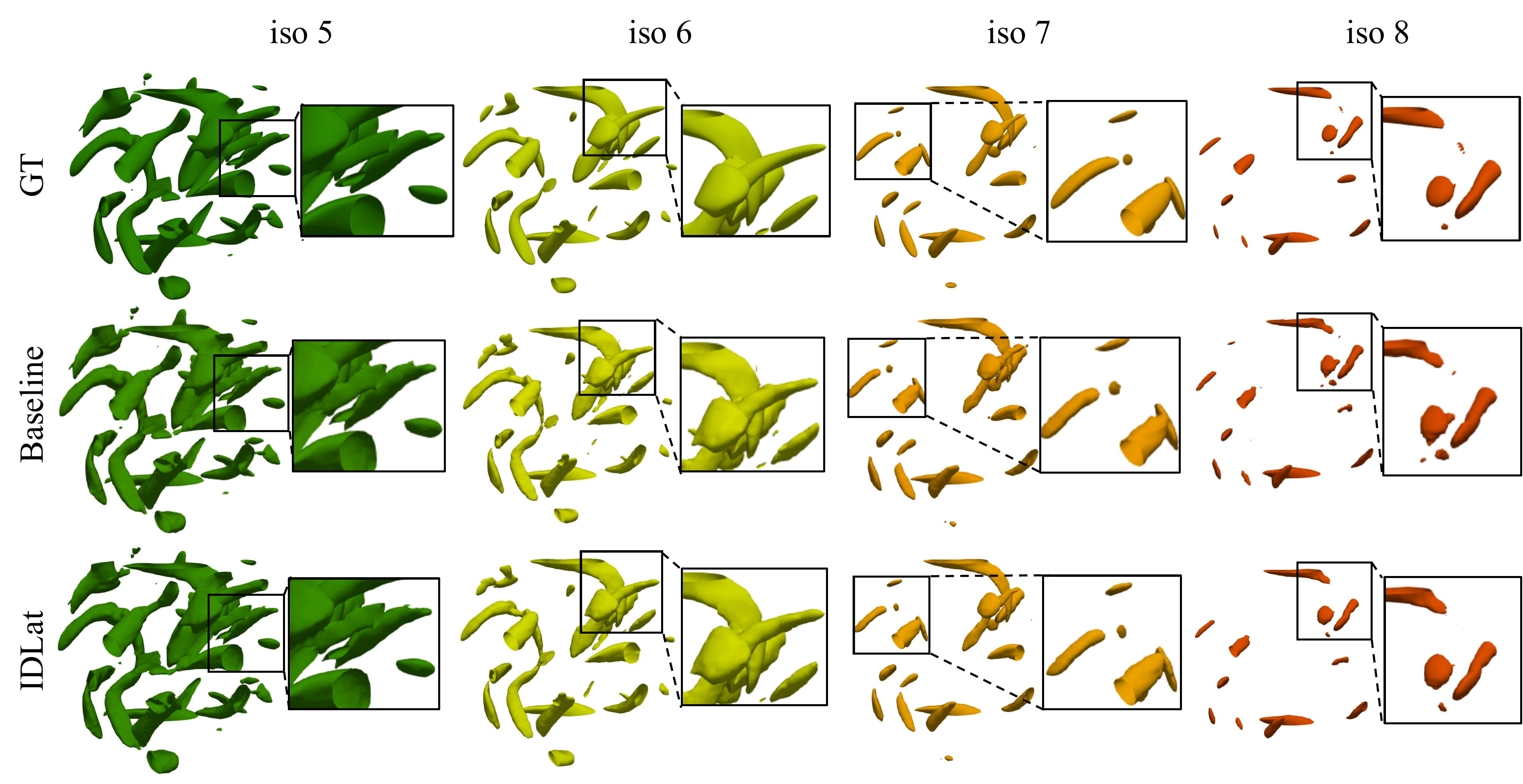}
    \vspace{-22pt}
    \caption{Comparison of isosurface rendering results of Vortex at time step 6 between ground truth (top row), baseline (middle row) and {\sysname}'s reconstruction (bottom row). The four importance maps for {\sysname} are defined based on distances to isosurfaces 5, 6, 7 and 8.}
    \label{fig:vortex_ts06_iso6}
\vspace{-11pt}
\end{figure}

\begin{figure*}[htp]
    \centering
    \includegraphics[width=18cm]{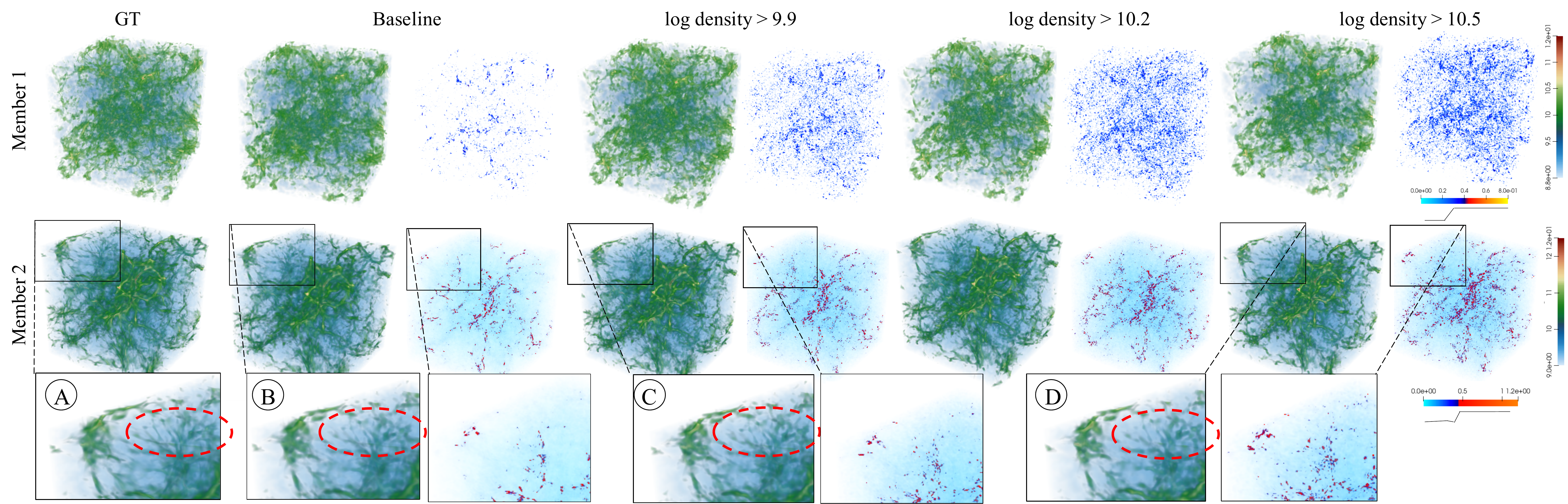}
    \vspace{-9pt}
    \caption{Volume rendering of ground truth, baseline and {\sysname}'s reconstructions on two ensemble members of Nyx data with value-based importance maps. Reference values are $9.9$, $10.2$ and $10.5$. The difference map for each reconstruction is shown on its right. 
    From left to right, in the difference map, the difference spreads out more when we increase the reference value, which matches the fact that as the reference value increases, the important regions are getting smaller and unimportant regions are enlarged so that the difference at unimportant regions becomes more obvious.}
    \label{fig:Nyx}
\vspace{-6pt}
\end{figure*}

\subsection{Qualitative Evaluation}
We qualitatively evaluate the proposed {\sysname} by visualizing the reconstructions with volume and isosurface rendering. 

In \autoref{fig:vortex_ts06_iso6}, we show isosurface rendering of the reconstructed Vortex data generated by the baseline and by {\sysname} with isosurface-distance-based importance maps.
In \autoref{fig:vortex_ts06_iso6}, in each row from left to right are isosurface rendering for isovalue 5, 6, 7, and 8.
Comparing isosurface rendering of {\sysname}'s reconstruction (third row) with the ground truth (first row) and the baseline (second row), we found that our latent representations can capture the structure and also the details in the ground truth, even though they are much smaller in size as discussed in~\autoref{sect:QuantiEval}. 
In some regions, {\sysname} can have slightly better reconstruction quality compared to the baseline, as shown in the zoom-in regions in ~\autoref{fig:vortex_ts06_iso6}.
These results demonstrate that our importance-driven latent representations can capture spatial importance information and have high reconstruction quality in important regions.

\autoref{fig:Nyx} displays volume rendering images of ground truth (first column), baseline, and {\sysname}'s reconstruction based on value-based importance maps for two ensemble members of the Nyx dataset.
The second and the third columns are volume rendering images of the baseline's reconstruction and difference maps between the original and the reconstructed data. Other columns are {\sysname}'s results based on three different value-based importance maps. The reference values are 9.9, 10.2, and 10.5, respectively.
In the first two rows, 
each row shows the result for one ensemble member. 
We use the same transfer function for each row to ensure the volume rendering difference is caused by the reconstructed data but not the transfer function difference. The transfer function for volume rendering of the reconstructed and ground truth data is the vertical colorbar and for difference map is the horizontal colorbar.
From volume rendering images, we found the latent representation has high reconstruction quality on every importance map. 
From left to right, we can see the differences in the difference map spreads out more when we increase the reference value, which matches the observation that as the reference value increases, the important regions in the dataset are getting smaller and unimportant regions are enlarged so that the difference at unimportant locations becomes more obvious. 
The third row of~\autoref{fig:Nyx} shows zoom-in of ground truth (A), baseline (B) and {\sysname}'s reconstruction (C and D). As shown in the figure, when the reference value increases from 9.9 in~\autoref{fig:Nyx} (C) to 10.5 in~\autoref{fig:Nyx} (D), the reconstruction loses more details of the unimportant data compared to the ground truth and the baseline, e.g., the red dashed circled regions are more smoothed out in \autoref{fig:Nyx} (D). The zoom-in of difference maps also shows more obvious spread-out differences in the unimportant regions. 
The pattern of difference maps and high-quality volume rendering results demonstrate that {\sysname} is under the guidance of spatial importance when generating latent representations. 

\begin{figure}[h]
    \centering
    \includegraphics[width=\columnwidth]{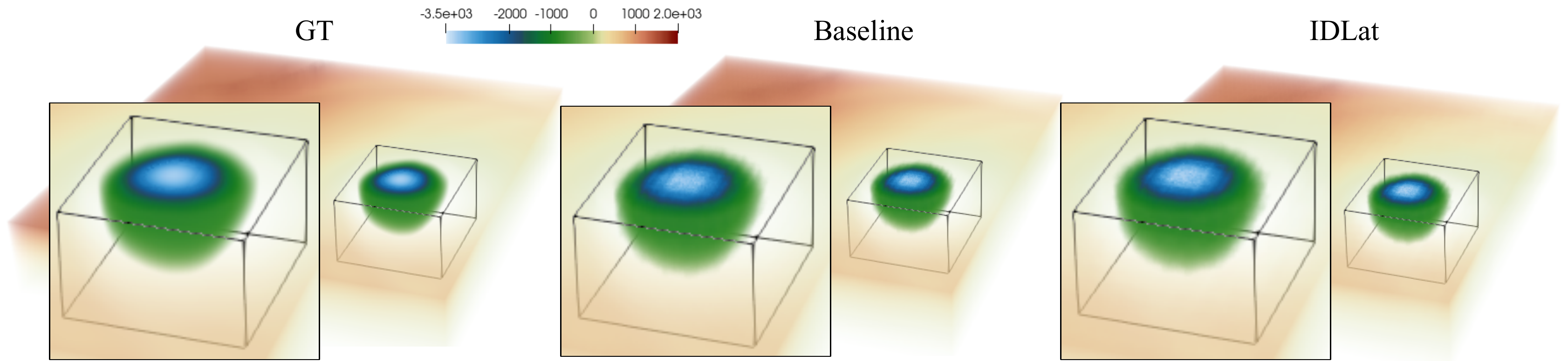}
    \vspace{-14pt}
    \caption{Volume rendering of ground truth, baseline and {\sysname}'s reconstruction on Isabel data with a location-based importance map where the hurricane eye in the bounding box is the region of interest.}
    \label{fig:isabel}
\vspace{-7pt}
\end{figure}

\autoref{fig:isabel} shows volume rendering images of Isabel dataset. From left to right, they are ground truth, baseline's, and {\sysname}’s reconstruction. 
{\sysname} utilizes a location-based importance map where the hurricane eye is the region of interest, as shown in the selected bounding box. Comparing the ground truth and baseline with {\sysname}'s reconstruction, we can see the quality of the hurricane eye is highly preserved, although {\sysname} has a smaller latent size as discussed in~\autoref{sect:QuantiEval}.

The above quantitative and qualitative results validate that {\sysname} is under spatial importance guidance when generating compact latent representations with high quality in important spatial regions.

\begin{figure}[h]
    \centering
    \includegraphics[width=\columnwidth]{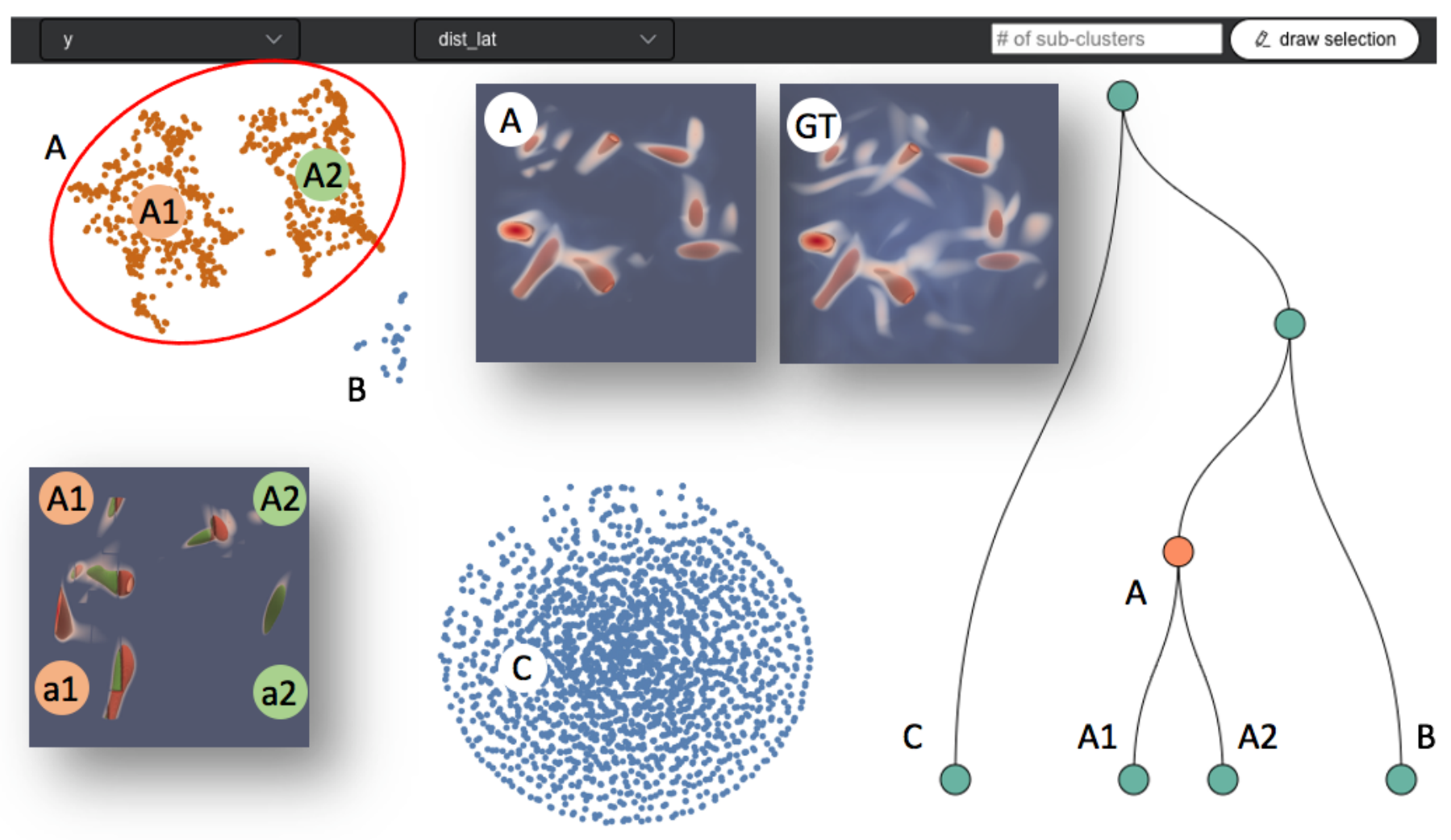}
    \vspace{-22pt}
    \caption{Latent space projection and  hierarchical clustering results given importance map based on distances to isosurface $8.0$. 
    Volume rendering and isosurface ($isovalue=8$) for the ground truth data (GT), for blocks from cluster A (A) and for child clusters of cluster A (bottom left), where isosurfaces are shown in red for cluster A1 and green for cluster A2.}
    \label{fig:iso8}
\vspace{-8pt}
\end{figure}

\subsection{Latent Space Exploration and Analysis}
As discussed in \autoref{sect:LSA}, our method produces latent representations that can be used for feature-related exploration and analysis. To show that the proposed importance-driven latent representations are succinct and suitable for representing features of interest, we perform latent space exploration and analysis on Vortex data with two case studies.

\subsubsection{Case Study 1: Obtain Insight of Features in Latent Space}
In the first case study, we show the effectiveness of importance-driven latents in identifying features by exploring the latent space. We compare block-wise latent representations generated by a uniform importance map and by a distance-based importance map in ~\autoref{sect:map_defination}.

First, we show the exploration results for the importance-driven latent space. The importance map is defined based on distances to the isosurface with $isovalue = 8$, which reveals the interest features of the vortex cores. In \autoref{fig:iso8}, we can easily identify four distinct clusters in the t-SNE projection of the latent vectors. The users can also modify the hierarchical clustering view to investigate the detail related to each cluster. 
On the right side of \autoref{fig:iso8}, we show the hierarchical clustering results of latent vectors. Cluster A consists of the blocks that contain the vortex cores and can be further separated into two child clusters (A1, A2) based on the separation in the t-SNE projection. The difference between cluster A1 and cluster A2 will be discussed later. Cluster C consists of blocks that do not intersect with the isosurface of interest. Cluster B contains some boundary blocks (e.g., blocks on the edges and in the corners).

To further investigate cluster A, we show images of volume rendering and isosurfaces for the ground truth data in \autoref{fig:iso8} (GT) and the blocks from cluster A identified in latent space in \autoref{fig:iso8} (A). It can be observed that \autoref{fig:iso8} (A) and \autoref{fig:iso8} (GT) reveal the same isosurfaces (ignoring the fuzzy region in the ground truth image, which is not part of the isosurface), indicating that our latent representations allow us to preserve the important regions with good quality and in the meantime enable us to visualize and separate the features easily in the latent space. 

\begin{figure}[htp]
    \centering
    \includegraphics[width=0.8\columnwidth]{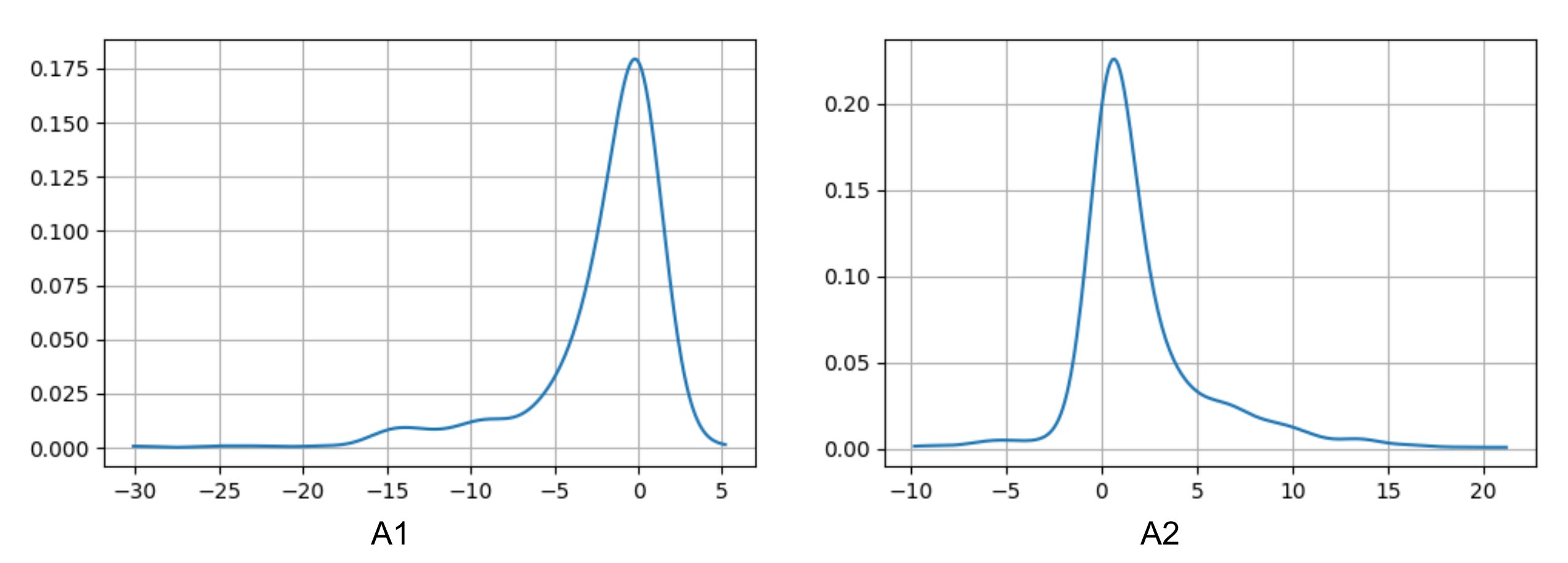}
    \vspace{-13pt}
    \caption{Gradient distribution for data blocks from two clusters.}
    \label{fig:grad}
\vspace{-12pt}
\end{figure}

In the t-SNE projection, cluster A consists of two child clusters A1 and A2. We show the isosurfaces corresponding to these two clusters in \autoref{fig:iso8} (A1, A2).  Cluster A1's isosurface is in red and cluster A2 in green. We found that except for two boundary vortices (a1, a2), all other vortices are split into two clusters (red and green). The splitting reveals the internal structures of the vortices. Since the splitting happens along one axis, we suspect one possible reason for this is the differences of the scalar values in the block along this direction. For this dataset, in the core of vortices, it has high scalar values and the value is decreasing and the isosurface is getting enlarged from inside vortex core to outside. They are classified into different clusters probably due to the opposite direction of value decreasing on this axis, which is related to the gradient of the values.
To validate this hypothesis, we calculate the average gradient distribution along the x-axis using Gaussian kernel density estimation for each data block from these two clusters as shown in \autoref{fig:grad}, where we can identify the apparent gradient distribution difference among these two clusters.
The further separation of the feature clusters helps visualize and understand the internal structures of the data of interest.

\begin{figure}[htp]
    \centering
    \includegraphics[width=0.55\columnwidth]{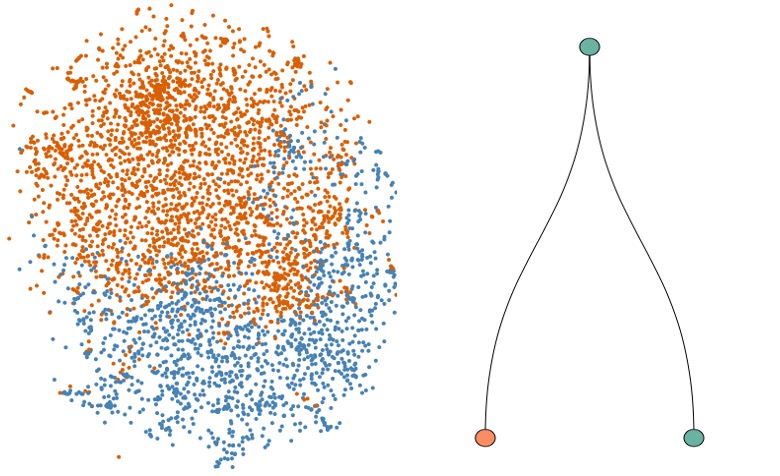}
    \vspace{-12pt}
    \caption{Latent exploration of latent vectors generated using uniform importance map. No visual clusters are identified in the t-SNE projection. }
    \label{fig:uni}
\vspace{-10pt}
\end{figure}

We also perform latent space exploration for latent vectors generated using a uniform importance map.
The t-SNE projection of their latent space is shown on the left side of \autoref{fig:uni}. We did not find any visual clusters in the t-SNE projection, and the clustering of the latent vectors splits all blocks into clusters of high and low average values, which is not helpful in feature-related analysis. The process of further splitting of the clusters is tedious and did not bring us anything interesting. 

By comparing the structures of importance-driven and uniform latent spaces, we found that the importance-driven latent space is highly related to the features of interest and is easier to explore. From the clustering result, scientists can reduce the effort of similarity comparison between blocks by quickly filtering out unimportant regions, resulting in fast and scalable data analysis.

\subsubsection{Case Study 2: Representative Isosurface Selection}
In the second case study, we show the usefulness of the importance-driven latent for representative isosurface selection.

To quantify and visualize similarities between isosurfaces, Bruckner and M{\"o}ller \cite{bruckner2010isosurface} proposed the isosurface similarity map. Each element in this map is the similarity value between two isosurfaces. They use distance fields, i.e., the minimal distance of each point to the surface, to represent isosurfaces and use mutual information between two distance fields as the similarity measure. However, one limitation of this method is the high computation cost. First, representing isosurfaces as distance fields is expensive without acceleration such as approximations. Second, generating the isosurface similarity map needs to calculate mutual information between every pair of isosurfaces, which requires building a joint histogram of every two distance fields. A more effective surface representation and efficient similarity computation is desired.

To solve this, we utilize {\sysname} with value-based importance maps to generate isosurface representations. We use value-based importance maps for two reasons. First, the spatial information of each voxel inside each block is encoded in the latent representation, so we do not need implicit distance fields to indicate surface locations. Second, when generating importance-driven latent representations, voxels with higher importance values will have a higher contribution, and voxels with low importance will be suppressed, 
which helps encode the surface information and zero out non-surface information. These two properties eliminate the heavy computation of distance fields. After we have latent representations for all blocks, we concatenate them into a single latent to represent the whole isosurface. We note that compared to the encoding time reported in~\autoref{table:dataset}, we have 8 times more blocks due to smaller block size ($8^3$ instead of $16^3$ as in~\autoref{table:dataset}), so representation generation is about 8 times slower. By changing the value-based importance maps, we can generate compact isosurface representations use different isovalues as the importance measure. Then, isosurface similarities are efficiently computed through cosine similarities between the isosurface-drive latent representations.

\begin{figure}[htp]
    \centering
    \includegraphics[width=0.8\columnwidth]{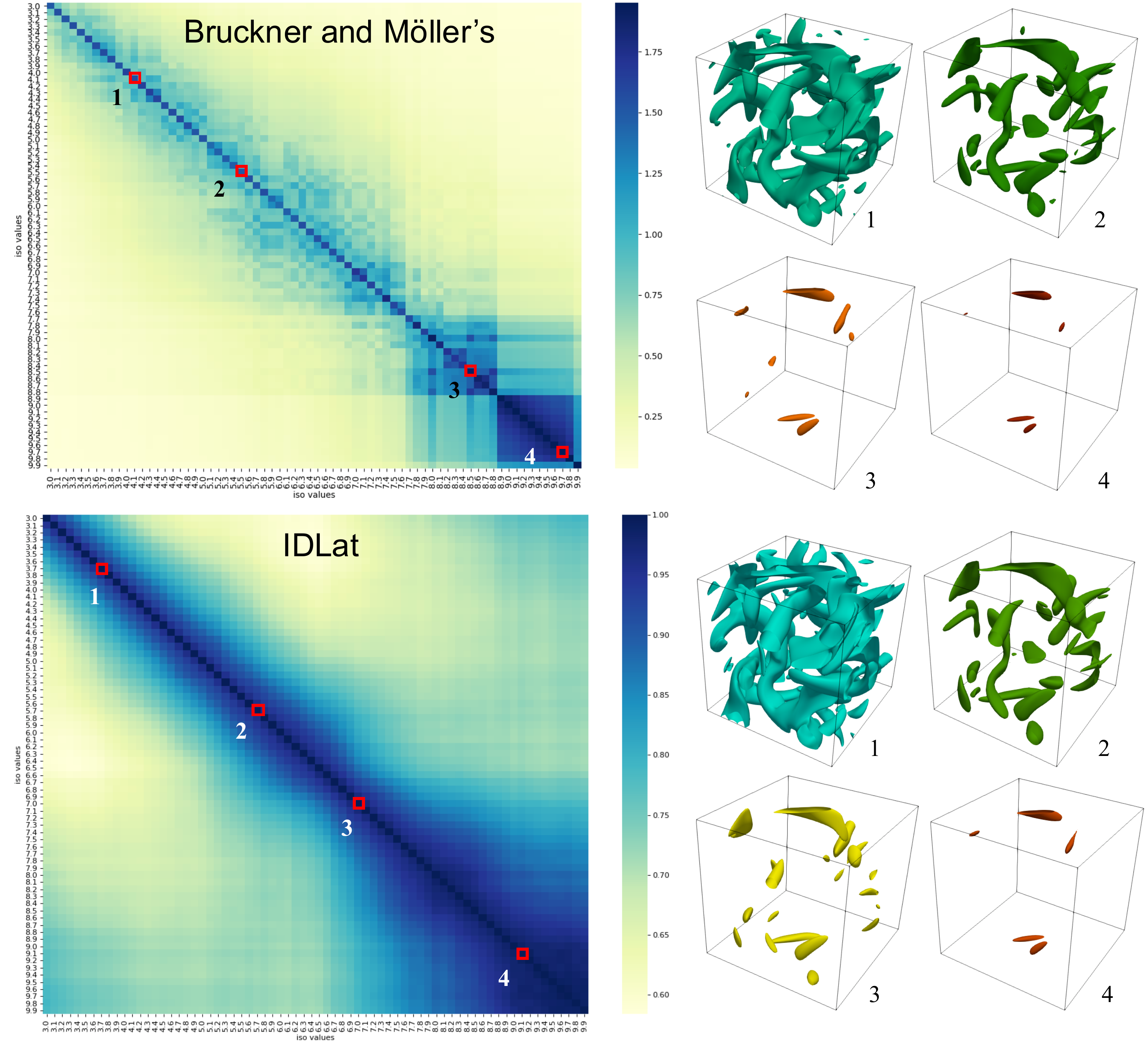}
    \vspace{-8pt}
    \caption{\textbf{Top Row:} Bruckner and M{\"o}ller's\cite{bruckner2010isosurface}
    isosurface similarity map (left) and four selected isosurfaces (right). \textbf{Bottom Row:} Our results of of isosurface similarity map (left) and four selected isosurfaces (right).}
    \label{fig:iso_selection}
\vspace{-4pt}
\end{figure}

In the isosurface similarity map, we can find clusters of isovalues to select representative isovalues. Given surface similarities, we use the same isosurface selection algorithm as Bruckner and M{\"o}ller's \cite{bruckner2010isosurface} to automatically identify representative isovalues.
\autoref{fig:iso_selection} shows isosurface similarity maps and selected isosurfaces computed by Bruckner and M{\"o}ller's method~\cite{bruckner2010isosurface} and by ours. 
Compared to Bruckner and M{\"o}ller's, our method can generate better results of the top four representative isosurfaces to reveal the structure of Vortex data. The selected isosurface (number 3) in our result is missing in theirs. They may identify it later but need to increase the number of selections.
In~\autoref{tab:iso_selection_time}, we show the performance of these two methods. Compared to Bruckner and M{\"o}ller's~\cite{bruckner2010isosurface}, our method is much more efficient in both representation generation stage (rep) and similarity computation stage (sim). 

\vspace{-5pt}
\begin{table}[htp]
\small
\caption{Selected isovalues, time (seconds) for all isosurface representation generation and for computing the isosurface similarity map using Bruckner and M{\"o}ller's \cite{bruckner2010isosurface} and importance-driven latent representations.}
\centering
 \begin{tabular}{p{25mm}|p{20mm}|p{11mm}|p{11mm}}
  Method & Selected Isovalues & Time (rep) & Time (sim)  \\ [0.5ex] 
  \hline
   Bruckner and M{\"o}ller's\cite{bruckner2010isosurface} & 4.1, 5.5, 8.5, 9.7 & 523.0288 & 824.9641 \\
  IDLat & 3.7, 5.7, 7.0, 9.1 & 183.8828 & 0.4095 
  \end{tabular}
  \label{tab:iso_selection_time}
\vspace{-10pt}
\end{table}

\section{Discussion and Future work} 
Even though we have demonstrated that the proposed {\sysname} can generate latent representations of compact size and correspond well to the importance definition, there are still several limitations to our work.

First, the generalizability of our method has not been fully investigated. Our evaluation results demonstrate that a model trained using data blocks from several time steps can generalize well to other time steps and is sensitive to different importance maps. However, to what extent our model can generalize is not fully understood. 


Secondly, in the practical use of our method, the importance definition and dataset itself may not always be in the same resolution. For example, in our isosurface-based importance definition, importance is continuously defined, while scalar data are only defined in the grid points, which forces us to sample the importance field to match the data resolution. How we interpolate and sample the importance field or the dataset can largely influence the latent representation quality.

Finally, the full potential of latent representation for scientific data analysis has not been extensively studied in this work. For example, the usage of importance-driven latent vectors on time-varying data analysis and feature tracking is one of our future studies. 
\vspace{-3pt}





\section{Conclusion} 
In this paper, we present an importance-driven latent generation method ({\sysname}) based on an autoencoder model which tightly relates latent representations to specific data of interest, such as salient regions or features of interest. 
We represent data of interest by spatial importance maps and utilize the location-wise importance information to guide latent generation. 
With a trained model, scientists can flexibly define various importance criteria and obtain different latent representations. We further reduce the latent size through a lossless entropy coding model. In addition, we develop a visual exploration tool for latent space analysis and demonstrate the efficiency of identifying and analyzing feature regions with importance-driven latent representations.  
Through quantitative and qualitative evaluations, we validate the effectiveness of our importance-driven latent generation method in representing data under domain interests control. 

\acknowledgments{
This work is supported in part by US Department of Energy SciDAC program DE-SC0021360, National Science Foundation Division of Information and Intelligent Systems IIS-1955764, and National Science Foundation Office of Advanced Cyberinfrastructure OAC-2112606. 
This research was also supported by the Laboratory Directed Research and Development program of Los Alamos National Laboratory under project number 20200065DR (LA-UR-22-23024).
}

\bibliographystyle{abbrv-doi}

\bibliography{template}
\end{document}